\documentclass[runningheads]{llncs}
\usepackage{graphicx}

\usepackage{tikz}
\usepackage{comment}
\usepackage{amsmath,amssymb} 
\usepackage{color}

\usepackage[accsupp]{axessibility}  

\usepackage[pagebackref,breaklinks,colorlinks]{hyperref}

\def\ourmodel{PL-CFE}
\usepackage{bm}
\usepackage{diagbox}
\DeclareMathOperator*{\argmax}{arg\,max}

\usepackage{floatrow}
\usepackage{multirow}

\begin{document}
\pagestyle{headings}
\mainmatter
\def\ECCVSubNumber{3487}  

\title{Rethinking Clustering-Based Pseudo-Labeling for Unsupervised Meta-Learning} 

\titlerunning{Rethinking Clustering-Based Pseudo-Labeling for UML}
%
\author{Xingping Dong\inst{1}\orcidID{0000-0003-1613-9288} \and
Jianbing Shen\inst{2}\thanks{Corresponding author: \textit{Jianbing Shen}}\orcidID{0000-0003-1883-2086} \and
Ling Shao\inst{3}\orcidID{0000-0002-8264-6117}}
\authorrunning{X. Dong et al.}
%
\institute{Inception Institute of Artificial Intelligence, Abu Dhabi, UAE 
\and SKL-IOTSC, Computer and Information Science, University of Macau
\and Terminus Group, China \\
\email{\{xingping.dong, shenjianbingcg\}@gmail.com}, \email{ling.shao@ieee.org}\\
}
\maketitle

\begin{abstract}
The pioneering method for unsupervised meta-learning, CACTUs, is a clustering-based approach with pseudo-labeling. This approach is model-agnostic and can be combined with supervised algorithms to learn from unlabeled data. However, it often suffers from label inconsistency or limited diversity, which leads to poor performance. In this work, we prove that the core reason for this is lack of a clustering-friendly property in the embedding space. We address this by minimizing the inter- to intra-class similarity ratio to provide clustering-friendly embedding features, and validate our approach through comprehensive experiments. Note that, despite only utilizing a simple clustering algorithm (k-means) in our embedding space to obtain the pseudo-labels, we achieve significant improvement. Moreover, we adopt a progressive evaluation mechanism to obtain more diverse samples in order to further alleviate the limited diversity problem. Finally, our approach is also model-agnostic and can easily be integrated into existing supervised methods. To demonstrate its generalization ability, we integrate it into two representative algorithms: MAML and EP. The results on three main few-shot benchmarks clearly show that the proposed method achieves significant improvement compared to state-of-the-art models. Notably, our approach also outperforms the corresponding supervised method in two tasks.
The code and models are available at \url{https://github.com/xingpingdong/PL-CFE}.

\keywords{Meta-learning; Unsupervised learning; Clustering-friendly}
\end{abstract}

\section{Introduction}

Recently, few-shot learning has attracted increasing attention in the {\it machine learning} and {\it computer vision} communities~\cite{li2019learning,Peng_2019_ICCV,Dvornik_2019_ICCV,ravichandran2019few,rodriguez2020embedding}. It is also commonly used to evaluate meta-learning approaches~\cite{hinton1987using,schmidhuber1987evolutionary,finn2017model}. However, most of the existing literature focuses on the supervised few-shot classification task, which is built upon datasets with human-specified labels. Thus, most previous works cannot make use of the rapidly increasing amount of unlabeled data from, for example, the internet. 

\begin{figure*}[t]
	
	\centering
	\includegraphics[width = .99 \textwidth]{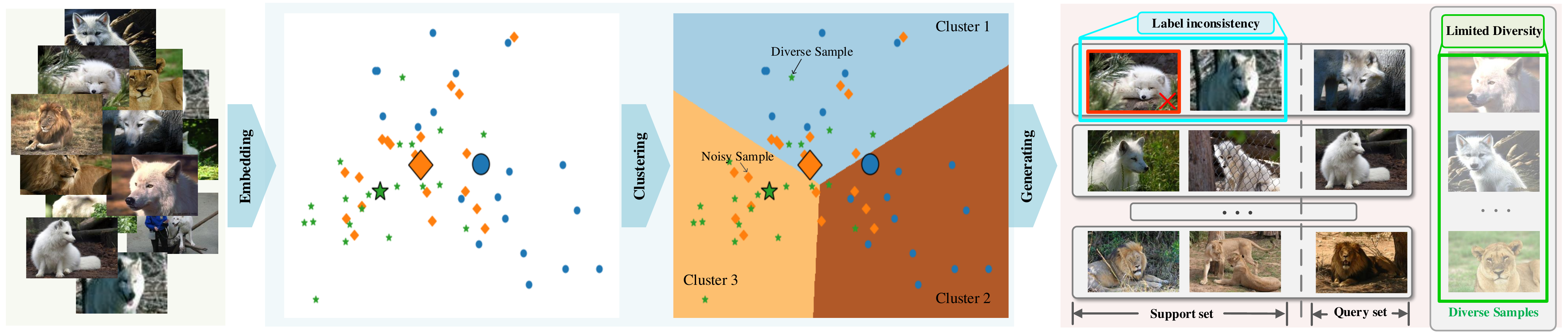}
	(a) Images~~~~~~~~~(b) PCA Space~~~~~~~~~~(c) Clusters ~~~~~~~~~~~(d) N-ways-K-shots task
	\caption{
		\textbf{Illustration of the \textit{label inconsistency} and \textit{limited diversity} issues in the clustering-based CACTUs~\cite{hsu2018unsupervised}.} \textbf{(a)} The unlabeled images. \textbf{(b)} The 2D mapping space of the embedding features, generated via principal component analysis (PCA). Each mark (or color) represents one class, and the larger marks are the class centers (\emph{i.e.} the average features in a class). \textbf{(c)} The noisy clustering labels generated by \textit{k-means}, with many inconsistent or noisy samples in each cluster. For example, we use the green class label for cluster 3, since cluster 3 has the most green-samples. Thus, the samples with inconsistent labels, like the orange samples in cluster 3, are regarded as noisy samples. Further, in the unsupervised setting, the green samples in other regions (cluster 1 and cluster 2) cannot be used to build few-shot tasks. This leads to the unsupervised few-shot tasks being less diverse than the supervised tasks. We term this issue \textit{limited diversity} and refer to these green samples as diverse samples. \textbf{(d)} The \textit{label inconsistency} and \textit{limited diversity} problems in the few-shot task, which are caused by the noisy clustering labels, as shown in (c). The red box highlights an incorrect sample in the few-shot task, and the green box illustrates the limited diversity problem, \emph{i.e.}, the diverse samples cannot be used to build few-shot tasks. 
		\vspace{-2mm}
	}
	\label{examples}
	\vspace{-2mm}
\end{figure*}

To solve this issue, the pioneering work CACTUs~\cite{hsu2018unsupervised}, was introduced to automatically construct the few-shot learning tasks by assigning pseudo-labels to the samples from unlabeled datasets. This approach partitions the samples into several clusters using a clustering algorithm ({\it k-means}) on their embedding features, which can be extracted by various unsupervised methods~\cite{hinton2006fast,bengio2007greedy,ranzato2007efficient,vincent2008extracting,erhan2010does}. Unlabeled samples in the same clusters are assigned the same pseudo-labels. Then, the supervised meta-learning methods, MAML~\cite{finn2017model} and ProtoNets~\cite{snell2017prototypical}, are applied to the few-shot tasks generated by this pseudo-labeled dataset. It is worth mentioning that CACTUs is model-agnostic and any other supervised methods can be used in this framework. 

However, this approach based on clustering and pseudo-label generation suffers from {\it label inconsistency}, {\emph{i.e.}} samples with the same pseudo-label may have different human-specified labels in the few-shot tasks (\textit{e.g.} Fig.~\ref{examples}(d)). This is caused by noisy clustering labels. 
Specifically, as shown in Fig.~\ref{examples}(c), several samples with different human-specified labels are partitioned into the same cluster, which leads to many noisy samples (with different labels) in the few-shot tasks. This is one reason for performance degeneration. Besides, noisy clustering labels will also partition samples with the same label into different clusters, which results in a lack of diversity for the few-shot tasks based on pseudo-labels compared with supervised methods. As shown in Fig.~\ref{examples} (c)(d), CACTUs ignores the diversity among partitioned samples with the same human label. This is termed the {\it limited diversity} problem. How to utilize the diversity is thus one critical issue for performance improvement.

To overcome the above problems, we first analyze the underlying reasons for the noisy clustering labels in depth, via a qualitative and quantitative comparison bewteen the embedding features of CACTUs. 
As shown in Fig.~\ref{examples}(b), we find that the embedding features extracted by unsupervised methods are not clustering-friendly. In other words, most samples are far away from their class centers, and different class centers are close to each other in the embedding space. Thus, a standard clustering algorithms cannot effectively partition samples from the same class into one cluster, leading to noisy clustering labels. For example, cluster 3 in Fig.~\ref{examples}(c) contains many noisy samples from different classes. Furthermore, we propose an inter- to intra-class similarity ratio to measure the clustering-friendly property. 
In the quantitative comparison, we observe that the accuracy of the few-shot task is inversely proportional to the similarity ratio. This indicates that reducing the similarity ratio is critical for performance improvement.

According to these observations, a novel pseudo-labeling framework based on a clustering-friendly feature embedding (\ourmodel) is proposed to construct few-shot tasks on unlabeled datasets, in order to alleviate the \textit{label inconsistency} and \textit{limited diversity} problems. Firstly, we introduce a new unsupervised training method to extract clustering-friendly embedding features. 
Since the similarity ratio can only be applied to labeled datasets, we simulate a labeled set via data augmentation and try to minimize the similarity ratio on this to provide a clustering-friendly embedding function. 
Given our embedding features, we can run \textit{k-means} to generate several clusters for pseudo-labeling and build clean few-shot tasks, reducing both the \textit{label inconsistency} and \textit{limited diversity} problems. Secondly, we present a progressive evaluation mechanism to obtain more divisive samples and further alleviate the \textit{limited diversity}, by utilizing additional clusters for the task construction. Specifically, for each cluster, which we call a base cluster, in the task construction, we choose its \textit{k-nearest} clusters as candidates. We use an evaluation model based on previous meta-learning models to measure the entropy of the candidates, and select the one with the highest entropy as the additional cluster for building a hard task, as it contains newer information for the current meta-learning model.

To evaluate the effectiveness of the proposed \ourmodel, we incorporate it into two representative supervised meta-learning methods: MAML~\cite{hsu2018unsupervised} and EP~\cite{rodriguez2020embedding}, termed as \ourmodel-MAML and \ourmodel-EP, respectively. We conduct extensive experiments on Omniglot~\cite{lake2011one}, \textit{mini}ImageNet~\cite{ravi2016optimization}, \textit{tiered}ImageNet \cite{ren2018meta}. The results demonstrate that our approach achieves significant improvement compared with state-of-the-art model-agnostic unsupervised meta-learning methods. In particular, our \ourmodel-MAML outperforms the corresponding supervised MAML method on \textit{mini}ImageNet in both the 5-ways-20-shots and 5-ways-50-shots tasks. Notably, we achieve a gain of 1.75\% in the latter task.

\section{Related Work}
\subsection{Meta-Learning for Few-Shot Classification}
{\it Meta-learning}, whose inception goes as far back as the   1980s~\cite{hinton1987using,schmidhuber1987evolutionary}, is usually interpolated as {\it fast weights}~\cite{hinton1987using,ba2016using}, or {\it learning-to-learn}~\cite{schmidhuber1987evolutionary,thrun1998learning,hochreiter2001learning,andrychowicz2016learning}. 
Recent meta-learning methods can be roughly split into three main categories. The first kind of approaches is metric-based~\cite{koch2015siamese,vinyals2016matching,snell2017prototypical,rodriguez2020embedding,chen2019closer}, which attempt to learn discriminative similarity metrics to distinguish samples from the same class. The second kind of approaches are memory-based~\cite{santoro2016meta,ravi2016optimization}, investigating the storing of key training examples with effective memory architectures or encoding fast adaptation methods. The last kind of approaches are optimization-based methods~\cite{finn2017model,finn2018probabilistic}, searching for adaptive initialization parameters, which can be quickly adjusted for new tasks. Most meta-learning methods are evaluated under supervised few-shot classification, which requires a large number of manual labels. In this paper, we explore a new approach of few-shot task construction for unsupervised meta-learning to reduce this requirement. 

\subsection{Unsupervised Meta-Learning}
{\it Unsupervised learning} aims to learn previously unknown patterns in a dataset without manual labels. These patterns learned during unsupervised pre-training can be used to more efficiently learn various downstream tasks, which is one of the more practical applications~\cite{hinton2006fast,bengio2007greedy,ranzato2007efficient,vincent2008extracting,erhan2010does}. Unsupervised pre-training has achieved significant success in several fields, such as image classification~\cite{zhang2017split,he2019momentum}, speech recognition~\cite{yu2010roles}, machine translation~\cite{ramachandran2016unsupervised}, and text classification~\cite{dai2015semi,howard2018universal,radford2018improving}.

Hsu {\it et al.}~\cite{hsu2018unsupervised} proposed an {\it unsupervised meta-learning method} based on clustering, named CACTUs, to explicitly extract effective patterns from small amounts of data for a variety of tasks. However, as mentioned, this method suffers from the \textit{label inconsistency} and \textit{limited diversity} problems, which our approach can significantly reduce. 
Subsequently, Khodadadeh {\it et al.}~\cite{khodadadeh2019unsupervised} proposed the UMTRA method to build synthetic training tasks 
in the meta-learning phase, by using random sampling and augmentation. 
Stemming from this, several works have been introduced to synthesize more generative and divisive meta-training tasks via various techniques, such as 
introducing a distribution shift between the support and query set~\cite{qin2020unsupervised}, using latent space interpolation in generative models to generate divisive samples~\cite{khodadadeh2020unsupervised}, and utilizing a variational autoencoder with a mixture of Gaussian for producing meta tasks and training~\cite{lee2020meta-gmvae}. Compared to these synthetic methods, our approach can obtain harder few-shot tasks.
Specifically, these previous methods only utilize the differences between augmented samples to increase the diversity of tasks, while our method can use the differences inside the class. 
{Besides, some researchers explore to apply clustering methods for meta-training, by prototypical transfer learning \cite{medina2020self} or progressive clustering \cite{ji2020unsupervised}. These methods are not model-agnostic, while our approach can be incorporated into any supervised meta-learning method. This will bridge the unsupervised and supervised methods in meta-learning. 
}

\begin{figure}[!tbh]
	\centering
	\includegraphics[width = .31\textwidth]{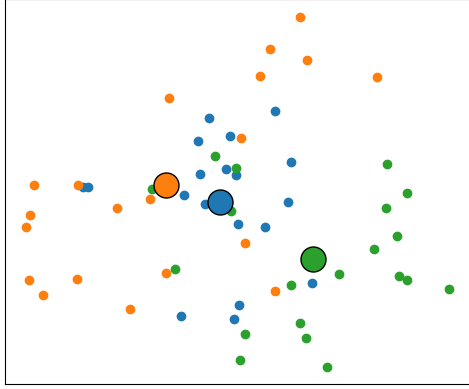}
	\includegraphics[width = .31\textwidth]{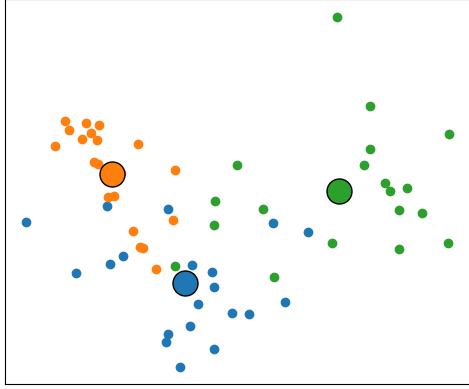}
	\includegraphics[width = .31\textwidth]{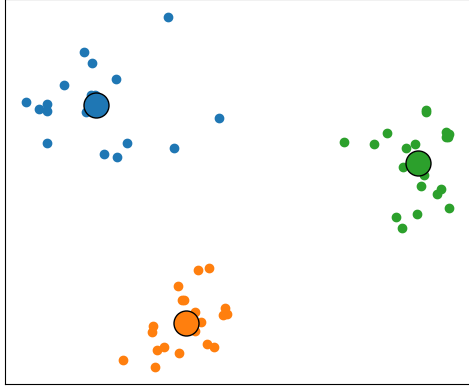}\\
	(a) BiGAN~~~~~~~~~~~~~~~~~~~~~(b) ACAI~~~~~~~~~~~~~~~~~~~~(c) CFE (ours)
	
	\caption{\textbf{Illustrations of 2D mapping of different embeddings, including BiGAN~\cite{donahue2016adversarial}, ACAI~\cite{berthelot2018understanding}, and our clustering-friendly embedding (CFE) features.} We randomly select three classes from Omniglot~\cite{lake2011one} and map the embeddings of their samples in 2D space via principal component analysis (PCA). Each color represents one class, and large circles are the class centers.
		\vspace{-2mm}
	}\label{fig:vis}
	\vspace{-2mm}
\end{figure}
\section{In-Depth Analysis of Clustering-Based Unsupervised Methods}

CACTUs~\cite{hsu2018unsupervised} is a clustering-based and model-agnostic algorithm, which is easily incorporated into supervised methods for the unsupervised meta-learning task. However, there is still a large gap in performance between CACTUs and the supervised methods. We believe that the main reason is that the unsupervised embedding algorithms, such as InfoGAN~\cite{chen2016infogan}, BiGAN~\cite{donahue2016adversarial}, ACAI~\cite{berthelot2018understanding} and DC~\cite{caron2018deep}, in CACTUs are not suitable for the clustering task, which is a core step of CACTUs. This is because these unsupervised methods were originally designed for the pre-training stage, where the extracted features can be further fine-tuned in the downstream tasks. Thus, they do not need to construct a clustering-friendly feature space, where the samples in the same class are close to their class center and each class center is far away from the samples in other classes. However, the clustering-friendly property is very important for CACTUs, since it directly clusters the embedding features without fine-tuning. 

We first provide an intuitive analysis by visualizing the embedding features. We collect samples from three randomly selected classes of a dataset and map them into 2D space with principal component analysis (PCA). For example, we observe the BiGAN and ACAI embedding features on Omniglot~\cite{lake2011one}. As shown in Fig.~\ref{fig:vis}(a)(b), many samples are far away from their class centers, and the class centers are close to each other. Thus, we can see that many samples in different classes are close to each other. These samples are difficult for a simple clustering method to partition. 

In order to observe the embedding features in the whole dataset, we define two metrics, \textit{intra-similarity} and \textit{inter-similarity}, to measure the clustering performance. We define the \textit{intra-similarity} for each class as follows:
\begin{equation}\label{eq:intra}
\small
s^{\text{intra}}_i = exp(\sum\nolimits_{j=1}^{N_s}{\bm\mu_{i} \cdot \mathbf{z}_{ij}} /(\tau N)),
\end{equation}
where $\mathbf{z}_{ij}$ is the embedding feature of the $j$-th sample in class $i$, $\bm\mu_i=\frac{1}{N_s}\sum\nolimits_{j=1}^{N_s}\mathbf{z}_{ij}$ is the class center, $\cdot$ is the dot product, $N_s$ is the number of samples in a class, and $\tau$ is a temperature hyperparameter \cite{wu2018unsupervised}. The \textit{intra-similarity} is the average similarity between the samples and their class center. It is used to evaluate the compactness of a class in the embedding space. A large value indicates that the class is compact and most samples are close to the class center. 
\begin{table}[!tb]
	\centering
	\setlength{\tabcolsep}{1.2mm}{
		\begin{tabular}{llcccc}
			\hline
			Embedding   & Dataset       & $s^{\text{inter}}$ \small$\downarrow$ & $s^{\text{intra}}$ \small$\uparrow$ & $R$ \small$\downarrow$   & Acc (\%) \small$\uparrow$\\
			\hline
			BIGAN~\cite{donahue2016adversarial}      & Omni      &\bf 1.032     & 2.690     & 0.449 & 58.18    \\
			ACAI~\cite{berthelot2018understanding}        & Omni      & 5.989     & 16.33    & 0.413 & 68.84    \\
			CFE (ours)   & Omni      & 1.316     &\bf 20.10    & \bf 0.081 &  \bf 91.32        \\
			\hline
			DC~\cite{caron2018deep} & Mini &\bf 0.999     & 1.164     & 0.862 & 39.90    \\
			CFE (ours)   & Mini & 1.048     &\bf 1.680     & \bf 0.641 &  \bf 41.99    \\
			\hline
		\end{tabular}
	}
	\caption{\textbf{Illustrations of the relationship between similarity ratio $R$ and classification accuracy (Acc).} We present the average \textit{intra-similarity} $s^{\text{intra}}$, \textit{inter-similarity} $s^{\text{inter}}$, and similarity ratio $R$ of different embedding methods on Omniglot~\cite{lake2011one}(Omni) and \textit{mini}ImageNet~\cite{ravi2016optimization}(Mini). We also report the accuracy of MAML~\cite{finn2017model} based on these embeddings for the 5-ways-1-shot task. All Acc values except ours are sourced from CACTUs~\cite{hsu2018unsupervised}.
		\vspace{-2mm}
	}\label{tab:simi-ratio}
	\vspace{-2mm}
\end{table}

The other metric, \textit{inter-similarity}, is defined as follows:
\begin{equation}\label{eq:inter}
\small
s^{\text{inter}}_{ij}=exp(\bm\mu_i\cdot \bm\mu_j / \tau),~j\neq i.
\end{equation}
A low value of $s^{\text{inter}}_{ij}$ indicates that two classes are far away from each other. 

To combine the above two similarities, we use the inter- to intra-class similarity ratio $r_{ij}=s^{\text{inter}}_{ij}/s^{\text{intra}}_i$ to represent the clustering performance. Finally, the average similarity ratio $R$ over the whole dataset is denoted as:
\begin{equation}\label{eq:R}
\small
R = {\frac{1}{C}\sum\nolimits_{i=1}^C\frac{\sum\nolimits_{j\neq i}^C s^{\text{inter}}_{ij}}{(C-1) s^{\text{intra}}_i}},
\end{equation}
where $C$ is the number of classes. The lower the value of $R$, the better the clustering performance. In addition to $R$, we also calculate the average \textit{intra-similarity} $s^{\text{intra}}=\sum{s^{\text{intra}}_i/C}$ and average \textit{inter-similarity} $s^{\text{inter}}=\sum{s^{\text{inter}}_{ij}/(N_sC)}$ for complete analysis, where $N_s$ is the samples’ number of a class.

As shown in Table~\ref{tab:simi-ratio}, we apply the above three criteria, $R$, $s^{\text{intra}}$ and $s^{\text{inter}}$, to the different embedding features (BiGAN~\cite{donahue2016adversarial} and ACAI~\cite{berthelot2018understanding}) in the Omniglot~\cite{lake2011one} dataset, and report the accuracy on the 5-ways-1-shot meta-task with the CACTUs-MAML~\cite{hsu2018unsupervised} method. We find that the accuracy is inversely proportional to the similarity ratio $R$, but has no explicit relationship with the individual similarities $s^{\text{intra}}$ or $s^{\text{inter}}$. This indicates that minimizing $R$ is critical for better accuracy.

Therefore, we propose a novel clustering-friendly embedding method (CFE) to reduce the similarity ratio in the following subsection \S\ref{sec:CF-embedding}. In the visualization comparison of Fig.~\ref{fig:vis}, the proposed method provides more compact classes than the previous BiGAN and ACAI. As shown in Table~\ref{tab:simi-ratio}, our CFE approach also has significantly reduced $R$ on both the Omniglot and \textit{mini}ImageNet. These results indicate that our method is more clustering-friendly. To investigate the relationship between the clustering-friendly property and the accuracy on the meta-task, 
we firstly use {\it k-means} to split the samples into different clusters,
and assign the same pseudo-label to samples in one cluster. Then we run the supervised MAML method on the 5-ways-1-shot task constructed by pseudo-labeled samples.
Compared with CACTUs, which is based on previous embedding methods, our approach achieves significant improvement on both Omniglot and \textit{mini}ImageNet. Notably, the gain on Omniglot is more than 20\%. The results clearly support our claim: a clustering-friendly embedding space can benefit clustering-based unsupervised meta-learning methods. 

\begin{figure*}[t]
	
	\centering
	\includegraphics[width = 0.99 \textwidth]{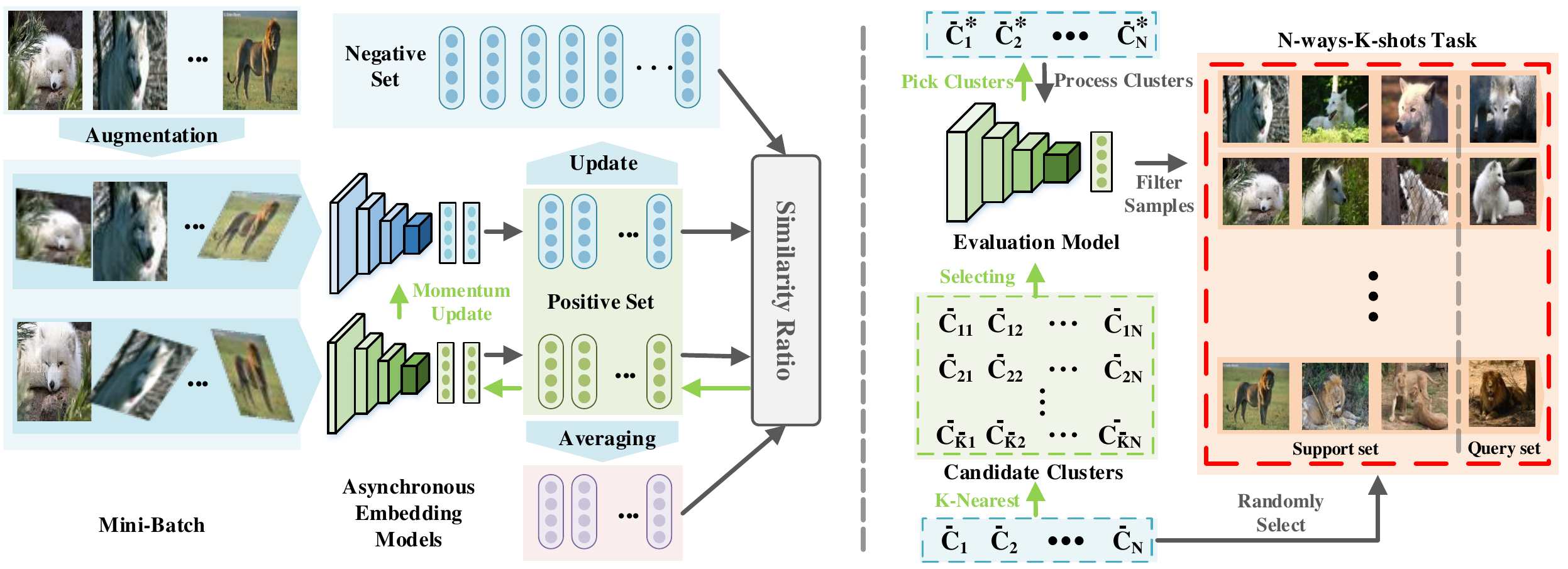}\\
	(a) Clustering-Friendly Feature Embedding  ~~~(b) Progressive Evaluation Mechanism
	\caption{
		\textbf{The frameworks of our clustering-friendly feature embedding and progressive evaluation mechanism.} \textbf{(a) The embedding feature extraction.} We first augment the samples in one mini-batch and feed them into the asynchronous embedding models to build a positive set. We also use the negative set storing the historical embedding features to obtain more varied features. Then we update our main embedding model (green) with backpropagation (green arrows), by minimizing the similarity ratio between the average features (purple ellipses), and the features in the positive and negative sets. The historical encoder is momentum updated with the main model. \textbf{(b) The progressive evaluation mechanism.} We randomly select $N$ clusters as base clusters ($\bar{C}_i$), and choose the $\bar{K}$-nearest neighbors as the candidates. Then, we use an evaluation model to select the clusters ($\bar{C}_i^*$) with the highest entropy and filter out noisy samples.
		\vspace{-2mm}
	}
	\vspace{-2mm}
	\label{framework}
\end{figure*}

\section{Our Approach}
We propose a novel clustering-based pseudo-labeling for unsupervised meta-learning, which includes a clustering-friendly feature embedding and a progressive evaluation mechanism.
Fig.~\ref{framework} shows the frameworks of our approach.
\subsection{Clustering-Friendly Feature Embedding}\label{sec:CF-embedding}
\noindent{\bf Optimization Objective.} We aim to learn a feature extraction function to map each sample to a clustering-friendly embedding space (with low similarity ratio). In this space, samples with the same class label are close to their class center, and each class center is far away from samples with different class labels. We can assign most samples from the same class into the same cluster, even with a simple clustering algorithm, such as \textit{k-means}. 

In the unsupervised setting, we do not have access to the class labels. Thus, we need to simulate a labeled dataset. To do so, we first randomly select $N_p\ll N_{all}$ samples from the unlabeled dataset $\mathcal{D}$ to build a positive set $\mathcal{D}_p$ , where $N_{all}$ is the size of $\mathcal{D}$. For each sample $\mathbf{x}_i\in\mathcal{D}_p$, we produce $N_a$ augmented samples $\{\mathbf{x}^+_{ij},j\in[1,N_a]\}$ via data augmentation methods, such as random color jittering, random horizontal flipping, and random grayscale conversion. Samples augmented from the same original sample are regarded as belonging to the same class. To involve more samples for training, we also construct a negative set by randomly selecting $N_n$ samples from $\mathcal{D}$, without including the $N_p$ positive samples. We augment each sample once to obtain the final negative set $\{\mathbf{x}^-_k,k\in[1,N_n]\}$. Given the positive (`labeled') set and negative set, we can reformulate the \textit{intra-similarity} in Eq.~(\ref{eq:intra}) and \textit{inter-similarity} in Eq.~(\ref{eq:inter}), for the final optimization objective. Maximizing the former will force the samples to be close to their class center, while minimizing the latter will pull the class centers far away from the other class center and negative samples.

We rewrite the \textit{intra-similarity} for each class:
\begin{equation}\label{eq:intra2}
\small
s^{\text{intra}}_i = exp(\sum\nolimits_{j=1}^{N_a}{\bm\mu_{i}\cdot \mathbf{z}^+_{ij}/(\tau N_a)}),
\end{equation}
where $\mathbf{z}^+_{ij}=\phi(\mathbf{x}^+_{ij};\bm\omega)$ is the embedding feature and $\bm\mu_i=\frac{1}{N_a}\sum\nolimits_j^{N_a}\mathbf{z}^+_{ij}$ is the class center. $\phi$ represents the embedding function and $\bm\omega$ is its parameter. This embedding function can be any learnable module, such as a convolutional network. 
The rewritten \textit{inter-similarity} includes a negative-similarity measuring the similarity between the class center and negative sample, and a center-similarity calculating the similarity bewteen the class centers. The formulations are as follows:
\begin{equation}\label{eq:inter2}
\small
s^{\text{neg}}_{ik}=exp(\bm\mu_i\cdot \mathbf{z}^-_{k}/\tau),~
s^{\text{cen}}_{ij}=exp(\bm\mu_i\cdot \bm\mu_j/\tau),
\end{equation}
where $\mathbf{z}^-_{k}=\phi(\mathbf{x}^-_{k};\bm\omega)$ is the negative embedding feature.

To utilize the common losses from standard deep learning tools, such as PyTorch~\cite{steiner2019pytorch}, we incorporate the logarithm function with the similarity ratio $R$ in Eq.~(\ref{eq:R}) to obtain the minimization objective, as follows:
\begin{equation}\label{eq:l1}
\small
L = \frac{1}{N_p}\sum\nolimits_{i=1}^{N_p}\log\frac{1}{N_o}(1+\frac{\sum\nolimits_{j\neq i}^{N_p} s^{\text{cen}}_{ij}+\sum\nolimits_{k=1}^{N_n}{s^{\text{neg}}_{k}}}{s^{\text{intra}}_i}),
\end{equation}
where $N_o = N_p+N_n-1$.
Similar to $R$ in Eq.~(\ref{eq:R}), a low value of $L$ in Eq.~(\ref{eq:l1}) means that the current embedding space is clustering-friendly. Thus, we can partition most samples from the same class into the same cluster, using a simple clustering algorithm. A low value of $L$ can also reduce the label inconsistency problem in the clustering-based unsupervised meta-learning. 
\noindent\textbf{Asynchronous Embedding.} If we only use a single encoder function $\phi$ for embedding learning, it is difficult to produce various embedding features for the samples augmented from the same original sample. Thus, we propose to utilize two asynchronous encoders, $\phi(\cdot;\bm\omega)$ and $\bar{\phi}(\cdot;\bar{\bm\omega})$ for training to avoid fast convergence caused by similar positive samples. The first function $\phi$ is the main encoder, which is updated with the current samples (in a mini-batch). In other words, we use gradient backpropagation to update its parameter $\bm\omega$. The latter $\bar{\phi}$ is the history encoder, which collects information from the main encoders in previous mini-batches. To ensure that the history encoders on-the-fly are smooth, \emph{i.e.}, the difference among these encoders can be made small~\cite{he2019momentum}, we use a momentum coefficient $m\in [0,1)$ to update the encoder parameter $\bar{\bm\omega} = m\bar{\bm\omega}+\bm\omega$.

To reduce the computational load, only one sample from each class is encoded by the main encoder $\phi$, and the others are encoded by the history encoder $\bar{\phi}$. Without loss of generality, we encode the first augmented sample in each class. Then, the embedding features of the positive dataset $\{\mathbf{x}^+_{ij},i\in[1,N_p],j\in[1,N_a]\}$ are reformulated as follows:
\begin{equation}
\small
\mathbf{z}^+_{ij}=
\begin{cases}
\phi(\mathbf{x}^+_{ij};\bm\omega),j=1,\\
\bar{\phi}(\mathbf{x}^+_{ij};\bar{\bm\omega}),j\neq 1.
\end{cases}
\end{equation}

For the negative set, a naive approach would be to randomly select samples from the unlabeled dataset and then encode them with the history encoder. However, this solution is time-consuming and does not make full use of the different history encoders from previous mini-batches. Inspired by \cite{he2019momentum}, we use a queue to construct the negative set by inserting the embedding features encoded by the history encoders. Specifically, we only insert one historical embedding of each class into the queue to maintain the diversity of the negative set. We remove the features of the oldest mini-batch to maintain the size of the negative set, since the oldest features are most outdated and least inconsistent with the new ones. Using the queue mechanism can reduce the computational cost and also remove the limit on the mini-batch size. 

\noindent\textbf{Clustering and Meta-Learning.} With the above training strategy, we can provide embedding features for the samples in the unlabeled dataset. Similar to CACTUs~\cite{hsu2018unsupervised}, we can then apply a simple clustering algorithm, \emph{e.g} \textit{k-means},to these features to split the samples into different clusters, denoted as $\{\mathcal{C}_1,\mathcal{C}_2,\cdots,\mathcal{C}_{N_s}\}$. We assign the same pseudo-label to all samples in a cluster. Thus, we obtain the pseudo-labeled dataset $\{(\mathbf{x}_i,\bar{y}_i), \mathbf{x}_i\in\mathcal{D}\}$, where $\bar{y}_i$ is the pseudo-label obtained by the clusters, \emph{i.e.} $\bar{y}_i=k$ if $\mathbf{x}_i\in\mathcal{C}_k$. Then, we can use any supervised meta-learning algorithm on this dataset to learn an efficient model for the meta-tasks.

\subsection{Progressive Evaluation Mechanism}
Although our embedding approach yields promising improvements, it still suffers from the \textit{limited diversity} problem. In other words, we cannot generate meta-tasks using the divisive samples, which are far away from their class center, and easily assigned the wrong label by the simple clustering algorithm. However, the supervised methods can utilize these samples to generate hard meta-tasks to enhance the model's discrimination ability. 
Thus, we propose a progressive evaluation mechanism to produce similar hard meta-tasks with supervised methods. 

As mentioned before, our embedding approach will push samples close to their class centers. For each cluster, denoted as the base cluster, we assume that the divisive samples are in its $\bar{K}$-nearest-neighbors, which are denoted as the candidate clusters. Thus, we can try to use the samples in the candidate clusters to generate hard meta-tasks. To do so, we build an evaluation model using the meta-learning models in previous iterations to evaluate the candidates. To obtain a stable evaluation model, only the meta-training models trained at the end of an epoch are utilized. We denote the evaluation model as ${f}(\mathbf{x};\mathbf{\bar{\bm\theta}})$, where $\mathbf{\bar{\bm\theta}}$ is the corresponding model parameter. Then, we select those with high information entropy as the final clusters, and choose the samples from the base and final clusters to generate hard meta-tasks, by filtering out noise. 



\noindent{\bf Selecting Clusters with High Information Entropy.} 
To construct an N-ways-K-shots meta-task, we need to randomly choose $N_c$ clusters denoted as base clusters $\bar{\mathcal{C}}_{i}$, $i\in[1,N_c]$. We use the cluster center similarity $\bar{s}_{ik}=\bar{\bm\mu}_i\cdot \bar{\bm\mu}_k$ to obtain the $\bar{K}$ most similar neighbors as the candidate clusters $\bar{\mathcal{C}}_{i\bar{k}}$, $k\in[1,\bar{K}]$, where $\bar{\bm\mu}_i$ is the average embedding feature of cluster $\bar{\mathcal{C}}_{i}$, and $\bar{K}=5$ in this paper. We randomly select K samples from each base cluster to build the support set, which we then use to finetune the evaluation model ${f}(\cdot ;\mathbf{\bar{\bm\theta}})$ with the same meta-training method. 

For simplification, we first define the entropy of a cluster $\mathcal{C}$ based on the finetuned evaluation model. The $N$-ways classification label of each sample $\mathbf{x}_i\in \mathcal{C}$ is computed as 
$l_i=\argmax\nolimits_{j=1}^{N}\mathbf{l}_i[j]$, where $\mathbf{l}_i=f(\mathbf{x}_i;\mathbf{\bar{\bm\theta}})$. Then we can provide the probability $p_j$ of selecting a sample with label $j$ from the cluster $\mathcal{C}$ by computing the frequency of the label $j$, \emph{i.e.}, $p_j = N_j/\dot{N}_l$, where $N_j$ is the occurrence number of label $j$ and $\dot{N}_l$ is the length of cluster $\mathcal{C}$. Then the entropy of $\mathcal{C}$ is formulated as: 
\begin{equation}
\label{eq:entropy}
\small
H(\mathcal{C})= -\sum\nolimits_{j=1}^{N}{p_j\log p_j}.
\end{equation}
According to Eq.~\ref{eq:entropy}, we choose the cluster $\mathcal{\bar{C}}_{i}^*$ with the highest entropy as the final cluster to construct the query set. The formulation is as follows: 
\begin{equation}
\small
\mathcal{\bar{C}}_{i}^* = \mathcal{\bar{C}}_{ik^*},  k^* = \arg\max\nolimits_{\bar{k}=1}^{\bar{K}} H(\mathcal{\bar{C}}_{i\bar{k}})
\end{equation}

In our case, a low entropy for a cluster indicates that it is certain or information-less for the evaluation model, since the label outcome of a cluster can be regarded as a variable based on our evaluation model. 
In other words, the information in a cluster with low entropy has already been learned in the previous training epochs and is thus not new for the current evaluation model. In contrast, a cluster with high entropy can provide unseen information for further improvement. 

\noindent{\bf Filtering Out Noisy Samples.} 
Notice that the cluster $\mathcal{\bar{C}}_{i}^*$ contains many noisy samples (with inconsistent labels compared with the support samples). Thus, we use the proposed evaluation model ${f}(\cdot;\mathbf{\bar{\bm\theta}})$ to filter out several noisy samples and build the query set.
First, we run ${f}(\cdot;\mathbf{\bar{\bm\theta}})$ on each sample $\mathbf{x}_{ij}$ in  $\mathcal{\bar{C}}_{i}^*$ to provide the {\it probabilities} $\mathbf{l}_{ij}$ of $N$-ways classification, \emph{i.e.}, $\mathbf{l}_{ij} = {f}(\mathbf{x}_{ij};\mathbf{\bar{\bm\theta}})$.
We take the {\it probability} of the $i$-th classification $\mathbf{l}_{ij}[i]$ as the evaluation score of the sample $\mathbf{x}_{ij}$. According to these scores, we re-sort the cluster $\mathcal{\bar{C}}_{i}^*$ in descending order to obtain a new cluster $\mathcal{\dot{C}}_{i}$. The noisy samples are placed at the end of the cluster. Thus, we can filter out the noisy samples by removing the ones at the end of the new cluster $\mathcal{\dot{C}}_{i}^*$. The {\it keep rate} $\beta\in(0,1)$ is used to control the number of samples removed. Specifically, we define the removing operation as $\mathcal{\dot{C}}_{i}=\mathcal{\dot{C}}_{i}[1:\lfloor \beta \bar{N}_l \rfloor]$, where $\bar{N}_l$ is the length of $\mathcal{\dot{C}}_{i}$, and $\lfloor\cdot\rfloor$ is the {\it floor} operation. Finally, we randomly select $Q$ samples from the cluster $\mathcal{\dot{C}}_{i}$ as the query set. 

In particular, during training, we employ a random value $\eta\in(0,1)$ for each mini-batch. Only when $\eta>0.9$, we use the progressive evaluation mechanism to build the meta-tasks. Since our pseudo-labels contain the most useful information for training, we need to utilize this information as much as possible.

\section{Experiments}\label{sec:exp}

\subsection{Datasets and Implementation Details}
\noindent{\bf Datasets.} We evaluate the proposed approach on three popular datasets: Omniglot~\cite{lake2011one}, \textit{mini}ImageNet~\cite{ravi2016optimization}, and \textit{tiered}ImageNet~\cite{ren2018meta}. Following the setting in \cite{hsu2018unsupervised}, for all three datasets, we only use the unlabeled data in the training subset to construct the unsupervised few-shot tasks.


\noindent\textbf{Embedding Functions and Hyperparameters.} For Omniglot, we adopt a 4-Conv model (the same as the one in MAML~\cite{finn2017model}), as the backbone, and add two fully connected (FC) layers with 64 output dimensions (64-FC). For \textit{mini}ImageNet and \textit{tiered}ImageNet, we use the same embedding function, which includes a ResNet-50~\cite{he2015delving} backbone and two 128-FC layers, and train it on the whole ImageNet. The other training details are the same for the two models. The number of training epochs is set to 200. We use the same data augmentation and cosine learning rate schedule as \cite{chen2020improved}. The other hyperparameters for training are set as $N_p=256$, $N_a=2$, $N_n=65536$, $\tau=0.2$, $m=0.999$. We use the same number of clusters as CACTUs for fair comparison, \emph{i.e.} $N_c=500$. The \textit{keep rate} $\beta$ in the progressive evaluation is set as 0.75 to filter out very noisy samples.

\begin{table}[!tbh]
\centering
	\resizebox{1.\textwidth}{!}{
		\setlength{\tabcolsep}{1mm}{
\begin{tabular}{cccccccccc|cccc}
\hline
                          &           &           &           &           &           & \multicolumn{4}{c}{Omniglot}    & \multicolumn{4}{c}{\textit{mini}ImageNet} \\
                          \hline
                          & Pos       & Neg       & AE        & SC        & FN        & (5,1) & (5,5) & (20,1) & (20,5) & (5,1) & (5,5) & (5,20) & (5,50) \\
                          \hline
A/D-M              &           &           &           &           &           & 68.84 & 87.78 & 48.09  & 73.36  & 39.90 & 53.97 & 63.84  & 69.64  \\
MC-M                 &           &           &           &           &           &  -     &    -   &     -   &    -    & 39.83 & 55.62 & 67.27  & 73.63  \\
\hline
\multirow{7}{*}{CFE-M} &\checkmark &           &           &           &           & 89.97 & 97.50 & 74.33  & 92.44  & 39.75 & 57.31 & 69.68  & 75.64  \\
                          &\checkmark &\checkmark &           &           &           & 69.88 & 89.68 & 40.50  & 72.50  & 25.35 & 34.91 & 47.80  & 55.92  \\
                          &\checkmark &           &\checkmark &           &           & 90.67 & 97.90 & 75.18  & 92.75  & 41.04 & 57.86 & 69.62  & 75.67  \\
                          &\checkmark &\checkmark &\checkmark &           &           & 91.32 & 97.96 & 76.19  & 93.30  & 41.99 & 58.66 & 69.64  & 75.46  \\
                          &\checkmark &\checkmark &\checkmark &\checkmark &           & 92.01 & 98.27 & 78.29  & 94.27  & 43.39 & 59.70 & 70.06  & 75.38  \\
                          &\checkmark &\checkmark &\checkmark &           &\checkmark & 91.40 & 97.94 & 78.61  & 94.46  & 42.85 & 59.31 & 69.61  & 75.25  \\
                          &\checkmark &\checkmark &\checkmark &\checkmark &\checkmark &\bf 92.81 &\bf 98.46 &\bf 80.86  &\bf 95.05  &\bf 43.60 &\bf 59.84 &\bf 71.19  &\bf 75.75  \\
                          \hline
                          \hline
                          
A/D-E                &           &           &           &           &           & 78.23 & 88.97 & 53.50  & 72.37  & 39.73 & 52.69 & 59.75  & 62.06  \\
MC-E                   &           &           &           &           &           &    -   &    -   &    -    &    -    & 43.26      & 56.67      &   63.19     &  65.07      \\
\hline
\multirow{7}{*}{CFE-E}   &\checkmark &           &           &           &           & 93.31 & 97.31 & 78.59  & 90.09  & 43.44 & 56.98 & 64.06  & 66.65  \\
                          &\checkmark &\checkmark &           &           &           & 60.70 & 73.37 & 37.93  & 49.33  & 26.68 & 32.36 & 37.17  & 38.86  \\
                          &\checkmark &           &\checkmark &           &           & 93.48 & 97.40 & 79.75  & 90.67  & 43.55 & 57.73 & 64.17  & 66.17  \\
                          &\checkmark &\checkmark &\checkmark &           &           & 93.88 & 97.46 & 79.55  & 91.13  & 47.55 & 62.13 & 68.89  & 71.11  \\
                          &\checkmark &\checkmark &\checkmark &\checkmark &           & 95.31 & 98.01 & 83.17  & 92.50  & 48.25 & 62.54 & 69.76  & 71.58  \\
                          &\checkmark &\checkmark &\checkmark &           &\checkmark & 94.93 & 98.00 & 82.57  & 91.97  & 48.02 & 62.56 & 69.85  &\bf 72.31  \\
                          &\checkmark &\checkmark &\checkmark &\checkmark &\checkmark &\bf 95.48 &\bf 98.04 &\bf 83.67  &\bf 92.54  &\bf 49.13 &\bf 62.91 &\bf 70.47  & 71.79 \\
                          \hline
\end{tabular}

}
}
\caption{\textbf{Influence of key components in our model.} We evaluate different variants in terms of the accuracy for N-ways-K-shots (N,K) tasks. The A/D labels represent ACAI~\cite{berthelot2018understanding} on Omniglot~\cite{lake2011one} and DC~\cite{caron2018deep} on \textit{mini}ImageNet~\cite{ravi2016optimization}. MC and M are short names of MoCo-v2~\cite{chen2020improved} and MAML~\cite{finn2017model}. Similarly, CFE is our embedding method and E represents the EP~\cite{rodriguez2020embedding}. The results of A/D-M are taken from CACTUs~\cite{hsu2018unsupervised}.
		\vspace{-2mm}
	}\label{tab:key components}
	\vspace{-2mm}
\end{table}


\noindent{\bf Supervised Meta-Learning Methods.} We combine the proposed pseudo-labeling based on clustering-friendly embedding (\ourmodel) with two supervised methods including classical model-agnostic meta-learning (MAML)~\cite{finn2017model}, and the recently proposed embedding propagation (EP)~\cite{rodriguez2020embedding}. The corresponding methods are denoted as \ourmodel-MAML and \ourmodel-EP, respectively. 
In the following experiments, we only train the meta-learning models (MAML and EP) on the one-shot task and directly test them on the other few-shot tasks. 

\subsection{Ablation Study}

To analyze the effectiveness of the key components in our framework, we conduct extensive experiments on Omniglot and \textit{mini}ImageNet with MAML and EP. First, we choose the unsupervised method ACAI~\cite{berthelot2018understanding}/DC~\cite{caron2018deep} as the baselines of the embedding function, which achieves the best accuracy for CACTUs. Specifically, we apply \textit{k-means} to the ACAI/DC embedding features to provide the pseudo-labels, and then run MAML and EP over eight few-shot tasks. These two baselines are denoted as A/D-M and A/D-E, respectively. The results of A/D-M come from the original CACTUs paper.

The key components in our clustering-friendly embedding (CFE) are the positive set (Pos), negative set (Neg), and asynchronous embedding (AE). We also investigate the main parts of our progressive evaluation, including selecting the cluster (SC) and filtering out noise (FN). First, we minimize the similarity ratio on the positive set. As shown in Table~\ref{tab:key components}, our approach achieves impressive improvement for both MAML and EP in terms of all four tasks on Omniglot, compared with the baselines. In particular, in the 5-ways-1-shot task, the highest accuracy gain reaches 21.15\%. Then, we add the negative set for training. However, the performance drops dramatically. The reason may be that the rapidly changing embedding models reduces the consistency of negative embedding features. Thus, we add AE to alleviate this issue. The increased performance (Pos+Neg+AE vs. Pos) indicates that Neg+AE can obtain more divisive samples for better training. We also use AE on the positive set (Pos+AE) and achieve expected accuracy gains for the two supervised methods in terms of all tasks. 
Finally, we use the SC strategy to select diverse samples for constructing few-shot tasks, yielding performance gains in all tasks. Then, the FN strategy is added to filter out noisy samples (SC+FN), which further improves the accuracy scores of all tasks. We also directly use our FN strategy on the original clusters (Pos+Neg+AE+FN) and achieve promising improvement in all tasks. This indicates that the proposed FN strategy can reduce the \textit{label inconsistency} issue for improved performance. 

Besides, we also compare the recent contrast learning method MoCo-v2~\cite{chen2020improved}, since our model has a similar training strategy but different training loss. We apply the pretrained MoCo-v2 model (200 epochs) to extract embeddings and provide the pseudo-labels, and then run MAML and EP over four few-shot tasks on \textit{mini}ImageNet. These are denoted as MC-M and MC-E, respectively.
As shown in Table~\ref{tab:key components}, even our methods without progress evaluation (Pos+Neg+AE) can outperform the MoCo-based methods in all tasks. This clearly demonstrates the effectiveness of our clustering-friendly embedding. Notice that our progress evaluation (Pos+Neg+AE+SC+FN) can further improve the performance.

\begin{table*}[!tb]
	\resizebox{0.99\textwidth}{!}{
		\begin{tabular}{lccccc|cccc}
			\hline
			&            & \multicolumn{4}{c}{Omniglot}    & \multicolumn{4}{c}{\textit{mini}ImageNet} \\
			\hline
			Algorithm (N, K)        & Clustering & (5,1) & (5,5) & (20,1) & (20,5) & (5,1)  & (5,5)  & (5,20) & (5,50) \\
			\hline
			Training from scratch~\cite{hsu2018unsupervised}   & -        & 52.50 & 74.78 & 24.91  & 47.62  & 27.59  & 38.48  & 51.53  & 59.63  \\
			\hline
            CACTUs-MAML~\cite{hsu2018unsupervised}             & BiGAN      & 58.18 & 78.66 & 35.56  & 58.62  & 36.24  & 51.28  & 61.33  & 66.91  \\
			CACTUs-ProtoNets~\cite{hsu2018unsupervised}        & BiGAN      & 54.74 & 71.69 & 33.40  & 50.62  & 36.62  & 50.16  & 59.56  & 63.27  \\
			CACTUs-MAML~\cite{hsu2018unsupervised}             & A/D  & 68.84 & 87.78 & 48.09  & 73.36  & 39.90  & 53.97  & 63.84  & 69.64  \\
			CACTUs-ProtoNets~\cite{hsu2018unsupervised}        & A/D  & 68.12 & 83.58 & 47.75  & 66.27  & 39.18  & 53.36  & 61.54  & 63.55  \\  
			AAL-ProtoNets~\cite{antoniou2019assume}            & -        & 84.66 & 89.14 & 68.79  & 74.28  & 37.67  & 40.29  &    -    &   -     \\
			AAL-MAML++~\cite{antoniou2019assume}              & -        & 88.40 & 97.96 & 70.21  & 88.32  & 34.57  & 49.18  &    -    &   -     \\
			ULDA-ProtoNets~\cite{qin2020unsupervised}           & -        &   91.00    &   98.14    &    78.05    &    94.08    & 40.63  & 56.18  & 64.31  & 66.43  \\
			ULDA-MetaOptNet~\cite{qin2020unsupervised}          & -        &   90.51    &   97.60    &    76.32    &   92.48     & 40.71  & 54.49  & 63.58  & 67.65  \\
			LASIUM-MAML~\cite{khodadadeh2020unsupervised}     & - & 83.26 & 95.29 &   -    &  -     & 40.19 & 54.56 & 65.17 & 69.13 \\
            LASIUM-ProtoNets~\cite{khodadadeh2020unsupervised} & - & 80.15 & 91.1  &    -   &   -    & 40.05 & 52.53 & 59.45 & 61.43 \\
			CACTUs-EP               & A/D  & 78.23 & 88.97 & 53.50  & 72.37  & 39.73  & 52.69  & 59.75  & 62.06  \\
			\ourmodel-MAML (ours)      & CFE        & 92.81 & \bf 98.46 & 80.86  & \bf 95.05  & 43.38  & 60.00  & \bf 70.64  & \bf 75.52  \\
			\ourmodel-EP (ours)       & CFE        & \bf 95.48 & 98.04 & \bf 83.67  & 92.54  & \bf 49.13  & \bf 62.91  & 70.47  & 71.79  \\
			\hline
			MAML (supervised)       & -        & 94.64 & 98.90 & 87.90  & 97.50  & 48.03  & 61.78  & 70.20  & 73.77  \\
			EP (supervised)         & -        & 98.24 & 99.23 & 92.38  & 97.18  & 57.15  & 71.27  & 77.46  & 78.77 \\
			\hline
		\end{tabular}
	}
	\small\caption{\textbf{Accuracy (\%) of N-ways-K-shots (N,K) tasks.} A/D represents ACAI~\cite{berthelot2018understanding} on Omniglot~\cite{lake2011one} and DC~\cite{caron2018deep} on \textit{mini}ImageNet~\cite{ravi2016optimization}. The best values are in bold. 
		\vspace{-2mm}
	}
	\label{tab:state-of-the}
	\vspace{-2mm}
\end{table*}

\subsection{Comparison with Other Algorithms}

\noindent\textbf{Results on Omniglot and \textit{mini}ImageNet.}
We compare our \ourmodel-MAML and \ourmodel-EP with model-agnostic methods: CACTUs~\cite{hsu2018unsupervised}, AAL~\cite{antoniou2019assume},   ULDA~\cite{qin2020unsupervised}, and LASIUM~\cite{khodadadeh2020unsupervised}.


As shown in Table~\ref{tab:state-of-the}, our \ourmodel-MAML outperforms all model-agnostic methods in eight few-shot tasks on two datasets. 
Specifically, for the 5-ways-5-shots task on the Omniglot dataset, our method achieves a very high score of $98.46\%$, which is very close to the supervised result of $98.90\%$. Surprisingly, the proposed \ourmodel-MAML even outperforms the corresponding supervised method under the 5-ways-20-shots and 5-ways-50-shots settings on the \textit{mini}ImageNet dataset. In addition, compared with the baseline CACTUs-MAML, we achieve significant accuracy gains, which reach 32.77\% (for 20-ways-1-shot) on Omniglot and 6.8\% (for 5-ways-20-shots) on \textit{mini}ImageNet. 

In the experiments for our \ourmodel-EP, we first create the CACTUs-EP baseline by combining CACTUs with EP. Our \ourmodel-EP provides impressive improvements in performance compared with CACTUs-EP. In particular, in the 20-ways-1-shot task on Omniglot, we achieve a huge gain of 30.17\%. On \textit{mini}ImageNet, the gains reach 10.72\% (in 5-ways-20-shots). Compared with the other existing methods, our \ourmodel-EP obtains superior performance in six tasks and comparable performance in the other two tasks. In particular, we achieve a significant accuracy gain of 8.42\% in 5-ways-1-shot on \textit{mini}ImageNet.   




\begin{table}[!tb]
\setlength{\tabcolsep}{1.2mm}{
\begin{tabular}{lcccc}
\hline
                      & (5,1) & (5,5) & (5,20) & (5,50) \\
                      \hline
Training from scratch \cite{qin2020unsupervised} & 26.27 & 34.91 & 38.14  & 38.67  \\
ULDA-ProtoNets \cite{qin2020unsupervised}       & 41.60 & 56.28 & 64.07  & 66.00  \\
ULDA-MetaOptNet \cite{qin2020unsupervised}       & 41.77 & 56.78 & 67.21  & 71.39  \\
PL-CFE-MAML (ours)          & 43.60 & 59.84 &\bf 71.19  &\bf 75.75       \\
PL-CFE-EP (ours)            &\bf 49.51 &\bf 64.31 & 70.98  & 73.06  \\
\hline
MAML (supervised) \cite{hsu2018unsupervised}    & 50.10 & 66.79 & 75.61  &  79.16  \\
EP (supervised) \cite{rodriguez2020embedding}       & 58.21 & 71.73 & 77.40  & 78.78 \\
\hline
\end{tabular}
}
\caption{\textbf{Accuracy (\%) of N-ways-K-shots (N,K) tasks on \textit{tiered}ImageNet~\cite{ren2018meta}.} We show the results of supervised methods MAML~\cite{hsu2018unsupervised} and EP~\cite{rodriguez2020embedding} for complete comparison. The best values are in bold.
\vspace{-2mm}
}
\vspace{-2mm}
\label{tab:tiered}
\end{table}

\noindent\textbf{Results on \textit{tiered}ImageNet.} 
We compare our \ourmodel-MAML and \ourmodel-EP with the recent ULDA~\cite{qin2020unsupervised} on \textit{tiered}ImageNet. As shown in Table~\ref{tab:tiered}, our models outperform the compared methods in four few-shot tasks. Specifically, for the 5-ways-1-shots and 5-ways-5-shots tasks, our \ourmodel-EP achieves the highest scores. Compared with the best previous method, ULDA-MetaOptNet, our method obtains accuracy gains of $7.74\%$ and $7.35\%$ on these two tasks, respectively. Our \ourmodel-MAML outperforms all compared methods in the 5-ways-20-shots and 5-ways-50-shots tasks. It also achieves significant improvements, with gains of $3.98\%$ and $4.36\%$, respectively, compared with ULDA-MetaOptNet. 

In summary, the results demonstrate the effectiveness of our method.

\section{Conclusion}
In this paper, we introduce a new framework of {pseudo-labeling based on clustering-friendly embedding} (\ourmodel) to automatically construct few-shot tasks from unlabeled datasets for meta-learning. 
Specifically, we present an unsupervised embedding approach to provide clustering-friendly features for few-shot tasks, which significantly reduces the {\it label inconsistency} and \textit{limited diversity} problems.
Moreover, a progressive evaluation is designed to build hard tasks to further alleviate \textit{limited diversity} issue. We successfully integrate the proposed method into two representative supervised models to demonstrate its generality.   
Finally, extensive empirical evaluations clearly demonstrate and the effectiveness of our \ourmodel, which outperforms the corresponding supervised meta-learning methods in two few-shot tasks. 
In the future, we will utilize our model to more computer vision tasks, such as object tracking \cite{dong2020clnet,han2021learning,shen2021distilled} and segmentation \cite{wu2022multi,wang2019inferring,dong2015sub}, to explore label-free or label-less solutions.
\clearpage
%
%
\bibliographystyle{splncs04}
\bibliography{egbib}
\end{document}